\pdfoutput=1
\documentclass[11pt]{article}

\usepackage[preprint]{acl}

\usepackage{times}
\usepackage{latexsym}
\usepackage{arydshln}
\usepackage{enumitem}
\usepackage[para]{footmisc}
\usepackage{array}
\usepackage{float}
\usepackage{booktabs} % For formal tables

\usepackage{booktabs}

\usepackage{multirow}
\usepackage{booktabs}

\usepackage[T1]{fontenc}
\usepackage[utf8]{inputenc}
\usepackage{amsmath}
\usepackage{cleveref}

\usepackage{microtype}

\usepackage{inconsolata}

\usepackage{graphicx}
\usepackage{adjustbox}  
\usepackage{array}  
\usepackage{hhline}

\usepackage{multirow}

\usepackage{todonotes}

\title{Change Is the Only Constant: Dynamic LLM Slicing based on Layer Redundancy } 

\author{
  \textbf{Razvan-Gabriel Dumitru\textsuperscript{1}},
  \textbf{Paul-Ioan Clotan\textsuperscript{2}},
  \textbf{Vikas Yadav\textsuperscript{3}},
  \textbf{Darius Peteleaza\textsuperscript{4, 5}},
\\
  \textbf{Mihai Surdeanu\textsuperscript{1}}
\\
  \textsuperscript{1}University of Arizona,
  \textsuperscript{2}Università di Bologna,
  \textsuperscript{3}ServiceNow AI, \\
  \textsuperscript{4}Lucian Blaga University of Sibiu, 
  \textsuperscript{5}MultiversX
\\
  \small{
    \textbf{Correspondence:} \href{razvandumm@gmail.com}{razvandumm@gmail.com}
  }
}

\begin{document}
\maketitle
\begin{abstract}

This paper introduces a novel model compression approach through dynamic layer-specific pruning in Large Language Models (LLMs), enhancing the traditional methodology established by SliceGPT. By transitioning from constant to dynamic slicing, our method leverages the newly proposed Layer Redundancy (LR) score, which assesses how much change each layer changes its input by measuring the cosine similarity of the input to the output of the layer. We use this score to prune parts of individual layers based on redundancy in such a way that the average pruned percentage for all layers is a fixed value. 
We conducted extensive experiments using models like Llama3-8B and Mistral-7B on multiple datasets, evaluating different slicing bases and percentages to determine optimal configurations that balance efficiency and performance. Our findings show that our dynamic slicing approach not only maintains but, in many cases, enhances model performance compared to the baseline established by constant slicing methods. For instance, in several settings, we see performance improvements of up to 5\% over the SliceGPT baseline.
Additionally, a perplexity decrease by as much as 7\% was observed across multiple benchmarks, validating the effectiveness of our method. The code, model weights, and datasets are open-sourced at \href{https://github.com/RazvanDu/DynamicSlicing}{https://github.com/RazvanDu/DynamicSlicing}. %\hoangshort{Having a short memorable name for the model would be great}

\end{abstract}

\section{Introduction}

Large Language Models (LLMs), characterized by their massive scale, often consist of billions to trillions of parameters, enabling them to perform a wide range of complex tasks with remarkable proficiency~\cite{kevian2024capabilities, llamapaper, team2023gemini, jiang2023mistral}. However, the deployment of these models poses significant challenges, primarily due to the extensive computational resources requirements. As the scale of these models grows, so does the urgency to develop more efficient methods for their deployment. This has led to increased interest in model compression techniques that aim to reduce the computational burden without substantially sacrificing performance. Techniques such as knowledge distillation, quantization, or pruning variants have emerged as viable solutions~\cite{wan2023efficient}, each offering a different approach to streamlining model architecture and operations \cite{wang2024model}. 

% \hoangshort{This paragraph is long and causes distractions for readers. Instead, I would suggest breaking it down to 2 paragraphs: (1) Introducing SliceGPT and its limitation, (2) Introducing the major contribution of the work. Start with "In our work, we... ".
% Suggestion as follows:
% SliceGPT[cite] is a model pruning technique via a constant slicing percentage of each layer. While.... network. (Add more shortcomings of SliceGPT if possible.

% Different from the ...., our work introduces a more nuanced,... Our proposed method aims to optimize both.... by preserving more functionality  }
%\hoangshort{I would recommend rewriting these paragraphs. Feel free to compare and choose the better one.}
In this paper, we improve the work on model pruning introduced by SliceGPT \cite{ashkboos2024slicegpt}, a pruning technique via a constant slicing percentage of each layer. While this approach reduces computational demands and maintains a level of performance, it does not account for the varying significance of different layers within the network.
We propose a more nuanced, dynamic pruning method that adapts the degree of pruning based on the individual characteristics and contributions of each layer. Our method aims to optimize both the efficiency and the efficacy of the pruning process by preserving more functionality in critical areas of the model, leading to better performance and less degradation in tasks.

%In this paper, we improve the work on model pruning introduced by SliceGPT \cite{ashkboos2024slicegpt}, a technique where layers within LLMs are pruned by slicing a constant percentage of each layer. While this approach reduces computational demands and maintains a level of performance, it does not account for the varying significance of different layers within the network. Addressing this limitation, our research proposes a more nuanced, dynamic pruning method that adapts the degree of pruning based on the individual characteristics and contributions of each layer. Different from the static, one-size-fits-all strategy of SliceGPT, this work aims to optimize both the efficiency and the efficacy of the pruning process. By not uniformly slicing across all layers, we seek to preserve more functionality in critical areas of the model, leading to better performance and less degradation in tasks.

%To quantify the impact of each layer on the model's overall performance, we developed a new metric called the Layer Redundancy (LR) score. 
%\hoangshort{I would not spend so much space on saying "similar to the ....". This does not help to understand/ emphasize your contribution. You could add this detail when introducing LR score in the method. }. 

%\hoangshort{Suggestions: Our results from the extensive empirical studies across various datasets and base models show} 
More specifically, we develop a new metric, namely Layer Redundancy (LR) score, to quantify  the impact of each layer on the model's overall performance. This evaluation is essential, as it guides the order in which layers are pruned, ensuring that the most influential layers are preserved while less critical layers are removed. Our approach involves generating slicing functions tailored to the importance of each layer, allowing for a dynamic and informed pruning strategy. Our results from the extensive empirical studies across various datasets and base models show a substantial improvement in model accuracy across all datasets tested, accompanied by a notable reduction in perplexity. In order to thoroughly evaluate the effectiveness of our proposed dynamic slicing pattern, we also analyzed the median accuracy and perplexity across a range of models. The results consistently show the superiority of our method over conventional constant slicing techniques.

The main contributions of our paper are:
\begin{itemize}[wide]
  \item We showed that dynamic, adaptive layer pruning can significantly improve computational efficiency without compromising model performance.
  \item The introduction of the Layer Redundancy score, a new metric to evaluate and guide the dynamic pruning of layers in LLMs.
  \item Extensive empirical validation shows that our method outperforms static pruning techniques in terms of both accuracy and perplexity across various settings.
\end{itemize}
\section{Related Work}

Recent advancements in model compression techniques \cite{hoefler2021sparsity, zhu2023survey} have markedly improved the efficiency of deploying LLMs while striving to retain their performance. The field has seen a variety of approaches including knowledge distillation 
\cite{hinton2015distilling,hsieh2023distilling}, quantization \cite{ma2024era}, pruning \cite{ma2023llm,yang2024laco}, low-rank adaptation \cite{hu2021lora} or hybrid variants \cite{xu2023qa,dettmers2024qlora}, each designed to address the growing computational and memory requirements of these models. 

Innovative approaches such as LLM-Pruner \cite{ma2023llm} and LaCo (Layer Collapse) \cite{yang2024laco} offer novel perspectives on model pruning. LLM-Pruner focuses on structured pruning by identifying and removing dependency groups within the model, aiming to minimize dependency on the original training corpus while preserving linguistic capabilities. Similarly, LaCo presents a layer-wise pruning strategy where subsequent layers collapse into preceding ones, achieving notable size reduction while maintaining good performance. A third approach \cite{gromov2024unreasonable} explores the potential of simple layer-pruning strategies combined with parameter-efficient finetuning (PEFT), demonstrating minimal performance loss even when half of the model's layers are removed. 

Among the innovative strategies in LLM optimization, SliceGPT \cite{ashkboos2024slicegpt} emerges as a significant breakthrough in model compression. Developed to address the intensive computational and memory demands of deploying LLMs, SliceGPT employs a unique post-training sparsification technique. Although effective in practice, previous research has illustrated that the order in which layers are removed plays a critical role in model performance \cite{gromov2024unreasonable,men2024shortgpt}. This insight led us to explore variable slicing percentages across different layers, challenging the constant slice for all layers. Initial attempts by the creators of SliceGPT to implement this through spectral analysis of layers did not yield a reliable method for determining the optimal percentage to be removed from each layer, as spectral analysis only tells part of the story and doesn't correlate with how much can be sliced out of a layer.

\section{Method}

\begin{figure*}[h]
\centering
\includegraphics[width=1.0\textwidth]{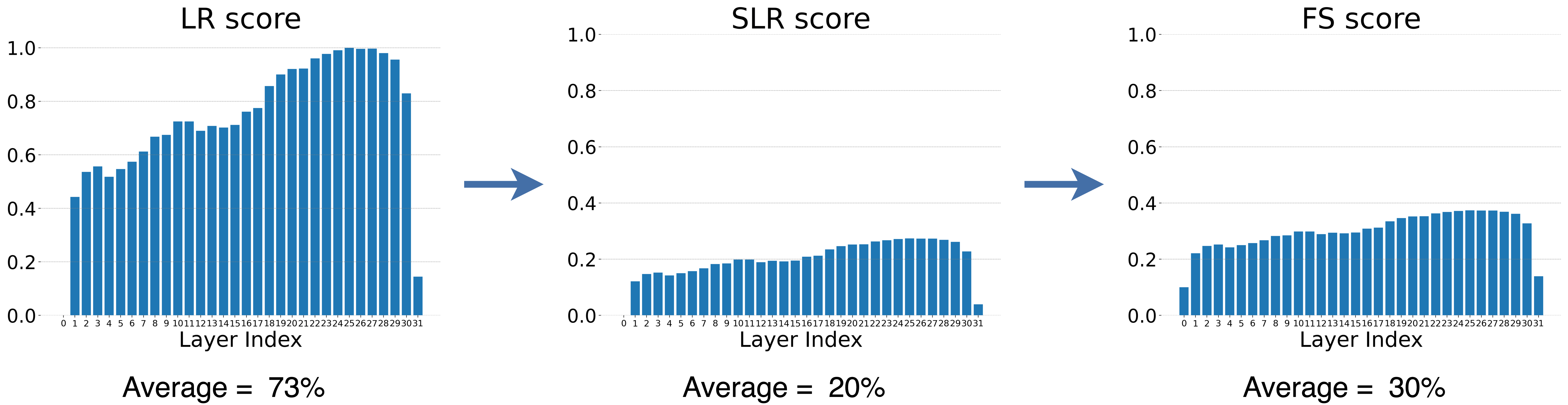}

% \vspace{-3mm}
\caption{Example of the Layer Redundancy (LR) score as well as the transformations used to achieve a slice percentage of 30\% ($S_P=0.3$) with a base slice for all layers of 10\% ($S_B=0.1$). The example is shown for the 32 layers of llama3-8B.}
\label{slicing_pattern}
% \vspace{-2mm}
\end{figure*}

%Our method is an improvement to the methodology defined in by SliceGPT \cite{ashkboos2024slicegpt} where layers within LLMs are pruned by slicing a constant percentage of each layer. While this works well in practice, it has been shown before that when it comes to removing layers completely, the ordering in which they are removed matters significantly \cite{gromov2024unreasonable}\cite{men2024shortgpt}. This idea prompted us to attempt to slice out different percentages of models, as opposed to a constant slice for all layers. The original SliceGPT paper attempted to do this by using a spectral analysis of the layers and they were unable to find a reliable way to decide what percentage should be sliced out of each layer.

\subsection{LR score}

Building on concepts introduced in a recent, unpublished study \cite{men2024shortgpt}, we propose a novel metric for assessing layer usefulness. This metric quantifies the extent to which each layer modifies its input by measuring the cosine similarity between the input and output. Specifically, we define this as the Layer Redundancy (LR) score:

% \vspace{-2mm}
\begin{equation}
    LR(\mathbf{L_i}) = \frac{\mathbf{L_i^I} \cdot \mathbf{L_i^O}}{\|\mathbf{L_i^I}\| \|\mathbf{L_i^O}\|}  
\label{LRScore}
\end{equation}

In Eq \ref{LRScore} $\mathbf{L_i^I}$ refers to the input of $layer_i$, while $\mathbf{L_i^O}$ refers to the output of $layer_i$. Intuitively, the higher cosine similarity between inputs and outputs leads to higher LR score, implying that the layer is more redundant. 
%\todox{You MUST explain what is novel here. At this point, the reader may think this is a clone of the ShortGPT paper... Is anything new in the LR score? Are you the first to use it for dynamic slicing?}
To evaluate the score for all layers, we use the full validation set of the PG-19 data set \cite{raecompressive2019}, we pass the data as context through the LLMs and we sum up how much each layer processed its input according to the cosine similarity score. At the end we normalize the values linearly so that the $min(LR(\mathbf{L_i}))=0$ and $max(LR(\mathbf{L_i}))=1$, guaranteeing that they range from 0 to 1. Intuitively, a LR score closer to 0 corresponds to lesser redundancy while higher LR score (close to 1) would mean high redundancy.
To the best of our knowledge, our work is the first one to use a per layer importance/redundancy score to prune variable parts of layers out (explained in the following section), instead of removing whole layers or removing a fixed constant portion of the layer \cite{men2024shortgpt}.

\subsection{Defining a per-layer percentage}
%\hoangshort{Suggestion: Based on the aforementioned LR score for each layer, we now seek to define the slicing mechanism for each layer such that .... The first step involves controlling the average of the $LR$ score which enables the ability to control the average pruned percentage of LLM. Concretely, we denote the average desired slicing percentage as $S_P$ (Slice Percentage) }

% Following passage was written by Dairus, I commented it out. See if this fits better.  
% Our goal is to have a function that slices out a fixed percentage of the LLM while having higher values for layers with high redundancy. To achieve this, we will apply several transformations on the score previously defined such that we have control over how big of an impact we want the redundancy to have on the sliced percentage.
Our goal is to have a function that slices variable sized parts of each layers based on their LR score while keeping the overall average of sliced out parts across all layers to be a fixed percentage of the LLM parameters specified by the user. For example, layers with higher LR values (or redundancy) can be sliced more compared to layers with lower LR score (or redundancy). To achieve this, we applied several transformations on the previously defined LR score such that we have control over how big of an impact we want the redundancy to have on the sliced percentage.
%\todox{This paragraph is backwards: you start with the solution before you introduce the idea. explain what your idea is first: slicing relative to an average fixed rate.}

The first step is to control the average of the $LR$ score so that we can then control the average pruned percentage of the LLM. Concretely, we denote the average desired slicing percentage as $S_P$ (Slice Percentage). In order to investigate what happens as we go further away from a constant slice for each layer, we will also define $S_B$ (Slice Base) to be a fixed constant value that will be guaranteed to be sliced from each layer such that $S_B <= S_P$. As $S_B=S_P$, the slicing becomes constant as presented by \citet{ashkboos2024slicegpt}. We experimented with different values of $S_B$ to see the effect of layer redundancy on the LLM's performance. 
%% A bit more clear version from Vikas
The next step is to scale the $LR$ function so that its average is $S_P-S_B$, thus once we add the base percentage for each layer ($S_B$) the total average will be $S_P-S_B+S_B=S_P$. This can be achieved by multiplying each $LR_{i}$ value by the ratio of $S_P-S_B$ divided by the mean of the function. This is denoted as Slice per Layer Redundancy ($SLR$) and shown in Eq \ref{FLR}.

% \vspace{-3mm}
\begin{equation}
    SLR(\mathbf{L_i}) = \text{LR}_i \cdot \frac{S_P - S_B}{\frac{1}{n} \sum_{i=1}^n \text{LR}_i}
\label{FLR}
\end{equation}

At the end, the Final Slice ($FS$) for the layer $\mathbf{L_i}$, is the sum of $SLR$ and $S_B$ as shown in Eq \ref{FC}. 
% \vspace{-1mm}
\begin{equation}
    FS(\mathbf{L_i}) = SLR(\mathbf{L_i}) + S_B
\label{FC}
% \vspace{-1mm}
\end{equation}

Average of $FS$ for all $n$ layers of a LLM is equal to $S_P$ as shown in \cref{AvgFSSPsame}.
% \vspace{-1mm}
\begin{equation}
    \frac{\sum_{i=1}^{n} FS(\mathbf{L_i})}{n} = S_P
\label{AvgFSSPsame}
\end{equation}

%% original written by Razvan/Darius
% The next step is to scale the $LR$ function so that its average is $S_P-S_B$, thus once we add the base percentage for each layer ($S_B$) the total average will be $S_P-S_B+S_B=S_P$. This can be achieved by multiplying each $LR_{i}$ value by the ratio of $S_P-S_B$ divided by the mean of the function. This is denoted as Fixed Layer Redundancy ($FLR$) and shown in Eq \ref{FLR}.

% \begin{equation}
%     FLR(\mathbf{L_i}) = \text{LR}_i \cdot \frac{S_P - S_B}{\frac{1}{n} \sum_{i=1}^n \text{LR}_i}
% \label{FLR}
% \end{equation}
% Finally, our slicing pattern, denoted as Final Slice ($FS$), is the sum of $FLR$ and $S_B$ as shown in Eq \ref{FC}. $FS$ a fixed average that is $S_P$.

% \begin{equation}
%     FS(\mathbf{L_i}) = FLR(\mathbf{L_i}) + S_B
% \label{FC}
% \end{equation}

\subsection{Slicing parts of layers}

\begin{table*}[h]
\centering
\footnotesize
% \resizebox{\textwidth}{!}{
\begin{tabular}{@{}ccc|cccccc@{}}
\toprule
\textbf{Model} & \textbf{Technique} & \textbf{Pruned} & \textbf{Piqa} & \textbf{Hellaswag} & \textbf{Winogrande} & \textbf{Arc Easy} & \textbf{Wikitextv2} & \textbf{Average} \\ 
& & & Acc. ($\uparrow$) & Acc. ($\uparrow$) & Acc. ($\uparrow$) & Acc. ($\uparrow$) & Perplexity ($\downarrow$) & Acc. ($\uparrow$) \\
\midrule
\multirow{6}{*}{\rotatebox[origin=c]{90}{{\bf Llama 3-8B}}} & \multirow{3}{*}{SliceGPT} & 30\% & 59.3\% & 37.2\% & 56.4\% & \textbf{42.9\%} & 13.37 & 49.0\% \\
                                 &                            & 35\%  & 57.7\% & 34.1\% & 54.3\% & 39.3\% & 16.58 & 46.4\% \\
                                 &                            & 40\%  & 57.0\% & 32.4\% & 51.8\% & 35.9\% & 20.69 & 44.3\% \\ \cmidrule{2-9}
                                 & \multirow{3}{*}{Dynamic Slicing} & 30\%  & \textbf{60.4\%} & \textbf{38.4\%} & \textbf{58.0\%} & 42.4\% & \textbf{12.96} & \textbf{49.8\%} \\
                                 &                            & 35\% & \textbf{58.4\%} & \textbf{36.3\%} & \textbf{57.2\%} & \textbf{39.3\%} & \textbf{15.64} & \textbf{47.8\%} \\
                                 &                            & 40\% & \textbf{58.1\%} & \textbf{34.0\%} & \textbf{54.4\%} & \textbf{36.8\%} & \textbf{19.11} & \textbf{45.8\%} \\
\cmidrule{2-9}
\multirow{6}{*}{\rotatebox[origin=c]{90}{{\bf Mistral-7B}}} & \multirow{3}{*}{SliceGPT} & 30\% & 62.6\% & 38.0\% & 59.7\% & 51.1\% & 8.87 & 52.9\% \\
                                 &                            & 35\% & 58.5\% & \textbf{35.9\%} & \textbf{57.6\%} & 42.8\% & 10.80 & 48.7\% \\
                                 &                            & 40\% & 57.1\% & \textbf{33.6\%} & 54.1\% & 38.2\% & 13.33 & 45.8\% \\ \cmidrule{2-9}
                                 & \multirow{3}{*}{Dynamic Slicing} & 30\% & \textbf{63.1\%} & \textbf{38.6\%} & \textbf{60.2\%} & \textbf{51.7\%} & \textbf{8.76} & \textbf{53.4\%} \\
                                 &                            & 35\% & \textbf{58.5\%} & 34.9\% & 55.7\% & \textbf{45.8\%} & \textbf{10.38} & \textbf{48.8\%} \\
                                 &                            & 40\% & \textbf{57.9\%} & 31.9\% & \textbf{54.1\%} & \textbf{40.1\%} & \textbf{12.62} & \textbf{46.0\%} \\
\bottomrule
\end{tabular}

\caption{Comparison of our technique in the smallest perplexity setting against the constant slicing proposed by SliceGPT; bold indicates higher values in comparison.}
\label{tab:techniques_comparison_slicegpt}
\end{table*}

\begin{figure*}[h]
% \vspace{-1mm}
\centering
\includegraphics[width=1.0\textwidth]{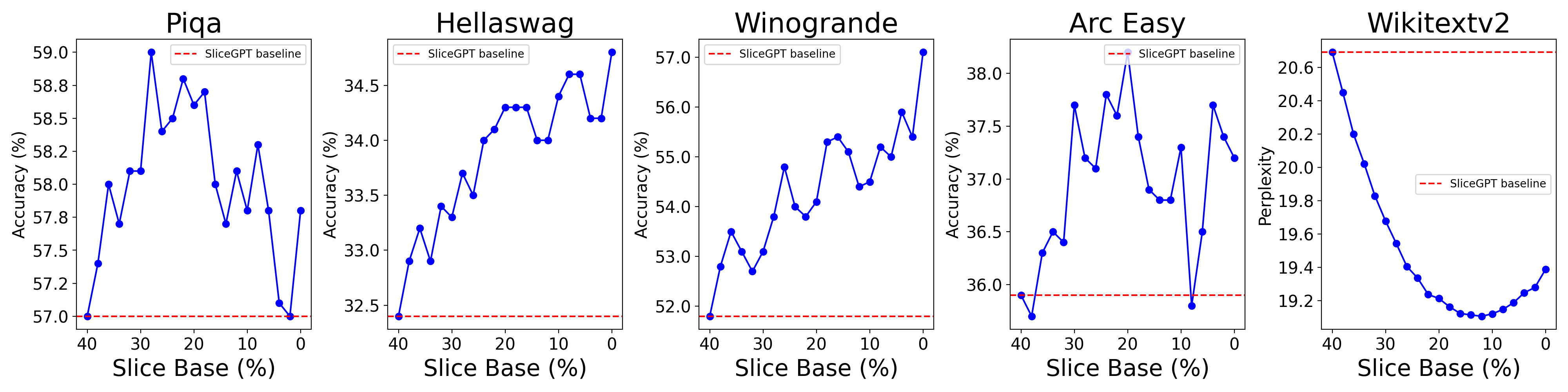}
% \vspace{-5mm}
\caption{Llama3-8B with 40\% of the network sliced on average, the red line is the baseline accuracy achieved by SliceGPT with a constant 40\% slice.}
\label{llama3b40p}
% \vspace{-2mm}
\end{figure*}

The methodology in \textit{SliceGPT} utilizes a specialized version of Principal Component Analysis (PCA) \cite{abdi2010principal} for efficient data reduction. It projects the data matrix $X$ onto a lower-dimensional subspace using the eigenvectors $Q$ and a deletion matrix $D$. The reduced matrix $Z$ is computed as $XQD$, where $D$ selectively omits certain components from $Q$, resulting in a compressed representation $Z$. The approximate reconstruction $\tilde{X}$ is obtained by $ZD^T Q^T$, minimizing the reconstruction error $\|\mathbf{X} - \mathbf{\tilde{X}}\|^2$. %\hoangshort{Need to define the high-level difference between SliceGPT and Ours. Suggestion: Unlike SliceGPT ...., we control } 
Unlike SliceGPT, we control the dimension of the matrix $D$ on a per-layer basis to achieve our dynamic slice.

%The methodology described in SliceGPT\cite{ashkboos2024slicegpt} employs a modified form of Principal Component Analysis (PCA), specifically adapted for efficient data representation reduction. Traditionally, PCA reduces the dimensionality of a data matrix $X$ by projecting it onto a lower-dimensional subspace using the eigenvectors $Q$ of $X^T X$, followed by a selection matrix $D$. In the paper, $Z$ is computed as $XQD$, where $D$ acts as a deletion matrix that selectively removes certain components (columns) from $Q$, effectively slicing the original matrix. This results in a compressed representation $Z$ of the original data, while the approximate reconstruction $\tilde{X}$ is given by $ZD^T Q^T$. This leverages the properties of PCA to maintain the essence of the data while minimizing the reconstruction error $\|\mathbf{X} - \mathbf{\tilde{X}}\|^2$.

%The deletion matrix $D$ is defined as a $D \times D_{small}$ deletion matrix (where $D_{\text{small}}$ columns of the $D \times D$ identity matrix are removed), which strategically removes specific rows and columns from the matrices involved in PCA. By applying $D$, only a subset of eigenvectors is retained in the computation, focusing on those that are deemed most critical for the task, thereby reducing the model's complexity. This selective reduction is crucial for maintaining computational efficiency, especially when processing large datasets. In our case, $FS(\mathbf{L_i})$ determines the size of the matrix $D_{small}$ that is used to remove part of $L_i$.
\section{Experiments}

The first step of the process is to evaluate how redundant each one of the layers is using the procedure described above. We have done this for Llama3-8B and Mistral-7B \cite{jiang2023mistral} using the full validation split of the pg-19 data-set \cite{raecompressive2019}. This will give us a $LR$ score for each layer that we then need to process into a slicing pattern. Figure \ref{slicing_pattern} shows an example of a slicing pattern evaluation process for Llama3-8B having a target $S_P$ of 30\% and a $S_B$ of 10\%.

Furthermore, we evaluate Llama3-8B and Mistral-7B using different Slice Base ($S_B$) values in increments of 2\%. Intuitively as we decrease the base percentage of the layers we have more extreme slicing patterns. In all our experiments we compare with the constant slice as a baseline. We experiment with Slice Percentages ($S_P$) of 30\%, 35\%, 40\%, these representing the percentage of the LLM that we prune. For evaluations, we use: Piqa \cite{bisk2019piqa}, Hellaswag \cite{zellers-etal-2019-hellaswag}, Winogrande \cite{DBLP:journals/corr/abs-1907-10641}, Arc Easy \cite{allenai:arc}, and Wikitextv2 \cite{merity2016pointer} within the library lm-evaluation-harness \cite{eval-harness}. For more experimental details, please refer to Section \ref{experimental_details_appendix} in the Appendix.

\section{Results}

\begin{table*}[h]
\centering
\footnotesize
% \resizebox{\textwidth}{!}{
\begin{tabular}{@{}ccc|cccccc@{}}
\toprule
\textbf{Model} & \textbf{Technique} & \textbf{Pruned} & \textbf{Piqa} & \textbf{Hellaswag} & \textbf{Winogrande} & \textbf{Arc Easy} & \textbf{Wikitextv2} & \textbf{Average} \\ 
& & & Acc. ($\uparrow$) & Acc. ($\uparrow$) & Acc. ($\uparrow$) & Acc. ($\uparrow$) & Perplexity ($\downarrow$) & Acc. ($\uparrow$) \\
\midrule
\multirow{6}{*}{\rotatebox[origin=c]{90}{\textbf{Llama 3-8B}}} & \multirow{3}{*}{ShortGPT} & 28.1\% & 62.0\% & 32.5\% & 57.5\% & 39.4\% & $2.2 \times 10^4$ & 47.9\% \\
                                 &                            & 34.3\% & 61.2\% & 34.3\% & 55.8\% & 38.1\% & $4.9 \times 10^4$ & 47.4\% \\
                                 &                            & 37.5\% & 60.0\% & 34.1\% & 54.8\% & 37.5\% & $1.1 \times 10^5$ & \textbf{46.6\%} \\ \cmidrule{2-9}
                                 & \multirow{3}{*}{Dynamic Slicing} & 30\% & 60.4\% & 38.4\% & 58.0\% & 42.4\% & $1.3 \times 10^1$ & \textbf{49.8\%} \\
                                 &                            & 35\% & 58.4\% & 36.3\% & 57.2\% & 39.3\% & $1.5 \times 10^1$ & \textbf{47.8\%} \\
                                 &                            & 40\% & 58.1\% & 34.0\% & 54.4\% & 36.8\% & $1.9 \times 10^1$ & 45.8\% \\
\cmidrule{2-9}
\multirow{6}{*}{\rotatebox[origin=c]{90}{\textbf{Mistral-7B}}} & \multirow{3}{*}{ShortGPT} & 28.1\% & 65.0\% & 39.8\% & 63.3\% & 46.6\% & $1.6 \times 10^2$ & \textbf{53.9\%} \\
                                 &                            & 34.3\% & 57.2\% & 28.6\% & 56.1\% & 30.3\% & $1.1 \times 10^4$ & 43.1\% \\
                                 &                            & 37.5\% & 55.6\% & 29.3\% & 56.7\% & 30.0\% & $2.4 \times 10^4$ & 42.9\% \\ \cmidrule{2-9}
                                 & \multirow{3}{*}{Dynamic Slicing} & 30\% & 63.1\% & 38.6\% & 60.2\% & 51.7\% & $8.7 \times 10^0$ & 53.4\% \\
                                 &                            & 35\% & 58.5\% & 34.9\% & 55.7\% & 45.8\% & $1.0 \times 10^1$ & \textbf{48.8\%} \\
                                 &                            & 40\% & 57.9\% & 31.9\% & 54.1\% & 40.1\% & $1.2 \times 10^1$ & \textbf{46.0\%} \\
\bottomrule
\end{tabular}
\caption{Comparison of our technique with ShortGPT. For each of the pruning ratios of our technique, the closest pruning ratio of ShortGPT is reported. Please note that removing 9, 11, and 12 layers completely as in ShortGPT results in 28.1\%, 34.3\%, and 37.5\% pruned ratio respectively. Bold indicates higher values in comparison.}
\label{tab:techniques_comparison}
%}
\end{table*}

We will first explore how the accuracy is affected by our dynamic slicing. Figure \ref{llama3b40p} (presented above),  and Figures \ref{llama3b30p}, \ref{llama3b35p}, \ref{mistralb30p}, \ref{mistralb35p}, \ref{mistralb40p} (presented in Appendix) show the behaviors of Llama3-8b and Mistral-7b on 30\%, 35\%, 40\% pruned percentage with respect to the accuracy on four data sets and perplexity on Wikitextv2. As observed in Figure \ref{llama3b40p}, for the Llama3-8B model, the accuracy is improved across all of the 5 evaluated datasets, and the perplexity decreases by as much as $1.4$ while pruning the exact same amount from the model. We also see huge accuracy improvements from 52\% to 57\% on Winogrande which has a baseline accuracy of 50\%. % Another point is that in all cases the perplexity decreased when we started to decrease the $S_B$, further showcasing that it improves the model's abilities.
An important finding in all cases is that the perplexity decreases and task accuracy (mostly) increases when we started to decrease the $S_B$ (until a certain point) showcasing LLMs benefit more from dynamic slicing as proposed in our work. 

% that LLMs benefit much more than variable slicing as proposed in our work compared to fixed/constant slicing. 
To estimate a good $S_B$ in our proposed dynamic slicing method, we evaluated the accuracy at the $S_B$ point that achieves the minimum perplexity on Wikitextv2, thus using it as a calibration data set. The results shown in Table \ref{tab:techniques_comparison_slicegpt} indicate that our method outperforms SliceGPT on average for all pruning ratios and models explored. We also show even better results using the median accuracy over all $S_B$ values (Table \ref{tab:median_comparison}) and mean accuracy values (Table \ref{tab:average_comparison}), leading us to believe that there are even better ways to choose an $S_B$ value than the minimum perplexity.

\subsection{Analyses}

To highlight strengths of pruning with our dynamic slicing, we also show comparison with ShortGPT~\cite{men2024shortgpt} where entire set of layers are pruned or removed. Please note that techniques such as ShortGPT that have shown to be effective, often provide lesser flexibility in pruning ratio as entire set of layers are removed. For example, removing 9 (least important) layers out of 32 layers results in a pruning ratio of 28.1\% as shown in Table \ref{tab:techniques_comparison}. 

As shown in Table \ref{tab:techniques_comparison}, our dynamic slicing with SliceGPT outperforms ShortGPT in majority of the cases even with higher pruning ration i.e., 28.1\% vs. 30\%, 34.3\% vs. 35\%, and 37.5\% vs. 40\%. Especially for higher pruning ratio (i.e., 35\% or 40\%), our dynamic slicing based SliceGPT approach outperforms ShortGPT with a larger margin across the four classification datasets shown in Table \ref{tab:techniques_comparison}.
Importantly, removing set of layers completely as in ShortGPT lead to very high perplexity values suggesting high degradation in text generation quality. On the other hand, our proposed dynamic slicing technique with SliceGPT results in exponentially better perplexity score highlighting benefits of pruning only parts of LLM layers instead of removing entire layers. 

Additionally, our variable slicing scaled from layer importance can be easily extended and merged with techniques like ShortGPT by simply removing the least important layers completely and slicing moderately important layers using our proposed approach. This hybrid method leveraging the strengths of both techniques, could potentially enhance model efficiency and performance. We leave this for exploration in future work.

\section{Conclusions}

%\todox{This paragraph is underwhelming. You should end strong. For example: We propose a novel dynamic slicing strategy that shows considerable improvements in accuracy when compared against a fixed slicing method that prunes the same amount of parameters on average...}
In conclusion, we propose a novel dynamic slicing strategy that shows considerable improvements in accuracy when compared against a fixed slicing method that prunes the same amount of parameters on average. We also show that layer redundancy is a powerful metric when removing percentages of layers, and also that there is room for improvement, leading to possible future work in the field.

%\todox{You don't need future work in a short paper. And you need the space. I recommend removing it.}

%Potential directions for future work are:

%\begin{itemize}
%    \item Finding better measures for layer importance or better way to transform importance into a slicing pattern.
%    \item Identifying the ideal $S_B$ that leads to the best improvement in accuracy across different tasks.
%    \item Exploring more models and settings in which the approach could be applied to.
%\end{itemize}

\section*{Limitations}

While our study introduces significant advancements in the dynamic pruning of Large Language Models, there are several limitations that are worth discussing:

\begin{itemize}
  \item The effectiveness of our method has been demonstrated predominantly on the Llama3-8B and Mistral-7B models. However, its performance may vary with other architectures, especially those with different layer configurations or learning dynamics. 
  \item We only experiment with one method to choose the $S_B$ value (that gives lowest perplixity) and there can be other methods for estimating $S_B$ which we leave it to more focused future works.
  \item Limited computational resources have constrained our ability to test on larger models. We believe that exploring the efficacy of our dynamic pruning method on more extensive architectures could provide valuable insights into its scalability and performance.
\end{itemize}

\section*{Ethical Considerations}

Ethical considerations are central to our development of a dynamic pruning method for large language models. Our research strives to reduce the computational costs and environmental impact of deploying large-scale models, aligning with the ethical responsibility to promote environmental sustainability and minimize negative consequences such as excessive energy consumption. This not only makes LLMs more accessible but also supports broader societal needs by enabling more efficient processing solutions that respect both individual rights and community values. By enhancing the efficiency of these models, we may enable populations with limited computational resources to harness the power of advanced NLP tools.

Furthermore, our methodology emphasizes the importance of using data sets that are free from harmful content. By using data sets that do not contain harmful data, we aim to ensure that the resulting models avoid biased outputs or content that could reduce their utility in practical applications. Ensuring the integrity and appropriateness of pruned models is essential, as these models often play significant roles in decision-making processes across various sectors. In this context, our approach is designed to be transparent and responsible, providing clear documentation and rigorous evaluation to maintain the reliability and fairness of the models.

%REMOVE AND SAY NO HARMFUL CONTENT IN DATA SETS. Furthermore, our methodology emphasizes the importance of avoiding harm. By refining the process of layer pruning, we aim to mitigate potential negative impacts on model performance that could result, for instance, in biased outputs or decreased utility in practical applications. Ensuring the integrity and usefulness of pruned models is essential, as these models often play significant roles in decision-making processes across various sectors. In this context, our approach is designed to be transparent and responsible, providing clear documentation and rigorous evaluation to support the reliability and fairness of the modified models. 

Also, we have selected a color scheme that prioritizes accessibility, ensuring the visuals are clear and discernible to individuals with color vision deficiencies. This inclusive approach reflects our commitment to making our research accessible to a wider audience, including those with varying visual abilities.

\bibliography{custom}

\appendix
\section{Appendix}

\subsection{Experimental details}
\label{experimental_details_appendix}

For all our experiments we used 4 NVIDIA A100 GPUs, with 80GB of VRAM each, and running all of the data sets on one slicing pattern using one NVIDIA A100 takes around 30 minutes, leading to a total time of 10 hours for a plot that has 20 possible $S_B$ values. We use a total of 1000 samples for each data set in all our configurations. Also, calculating the LR score can take upwards of 30-40 minutes per model on 1 NVIDIA A100 GPU, but that only has to be computed once. %\hoangshort{Push this to Appendix if you run out of space.}

\subsection{Tables with mean/median results}

For a detailed evaluation, Table \ref{tab:median_comparison} compares our technique in the median setting against the constant slicing proposed by SliceGPT, and Table \ref{tab:average_comparison} provides a comparison in the average setting.

\begin{table*}[h]
\centering
\resizebox{\textwidth}{!}{%
\begin{tabular}{@{}ccc|cccccc@{}}
\toprule
\textbf{Model} & \textbf{Technique} & \textbf{Pruned} & \textbf{Piqa} & \textbf{Hellaswag} & \textbf{Winogrande} & \textbf{Arc Easy} & \textbf{Wikitextv2} & \textbf{Average} \\ 
& & & Acc. ($\uparrow$) & Acc. ($\uparrow$) & Acc. ($\uparrow$) & Acc. ($\uparrow$) & Perplexity ($\downarrow$) & Acc. ($\uparrow$) \\
\midrule
\multirow{6}{*}{\rotatebox[origin=c]{90}{{\bf LLaMa-3-8B}}} & \multirow{3}{*}{SliceGPT} & 30\% & 59.3\% & 37.2\% & 56.4\% & 42.9\% & 13.37 & 49.0\% \\
&                            & 35\%  & 57.7\% & 34.1\% & 54.3\% & 39.3\% & 16.58 & 46.4\% \\
&                            & 40\%  & 57.0\% & 32.4\% & 51.8\% & 35.9\% & 20.69 & 44.3\% \\ \cmidrule{2-9}
& \multirow{3}{*}{Dynamic Slicing} & 30\%  & \textbf{60.5\%} & \textbf{38.1\%} & \textbf{57.4\%} & \textbf{43.0\%} & \textbf{13.07} & \textbf{49.7\%} \\
&                            & 35\% & \textbf{58.7\%} & \textbf{35.8\%} & \textbf{56.2\%} & \textbf{39.6\%} & \textbf{15.75} & \textbf{47.6\%} \\
&                            & 40\% & \textbf{58.1\%} & \textbf{34.2\%} & \textbf{55.0\%} & \textbf{37.2\%} & \textbf{19.21} & \textbf{46.1\%} \\
\cmidrule{2-9}
\multirow{6}{*}{\rotatebox[origin=c]{90}{{\bf Mistral-7B}}} & \multirow{3}{*}{SliceGPT} & 30\% & 62.6\% & 38.0\% & \textbf{59.7\%} & 51.1\% & 8.87 & 52.9\% \\
&                            & 35\% & 58.5\% & 35.9\% & \textbf{57.6\%} & 42.8\% & 10.80 & 48.7\% \\
&                            & 40\% & 57.1\% & 33.6\% & 54.1\% & 38.2\% & 13.33 & 45.8\% \\ \cmidrule{2-9}
& \multirow{3}{*}{Dynamic Slicing} & 30\% & \textbf{62.6\%} & \textbf{38.6\%} & 59.5\% & \textbf{52.4\%} & \textbf{8.80} & \textbf{53.4\%} \\
&                            & 35\% & \textbf{59.2\%} & \textbf{36.0\%} & 56.9\% & \textbf{45.1\%} & \textbf{10.42} & \textbf{49.3\%} \\
&                            & 40\% & \textbf{57.9\%} & \textbf{34.3\%} & \textbf{54.1\%} & \textbf{40.0\%} & \textbf{12.68} & \textbf{46.5\%} \\
\bottomrule
\end{tabular}
} % Correctly placed closing brace for \resizebox
\caption{Comparison of our technique in the median setting against the constant slicing proposed by SliceGPT; bold indicates higher values in comparison.}
\label{tab:median_comparison}
\end{table*}

\begin{table*}[h]
\centering
\resizebox{\textwidth}{!}{ % Removed the % symbol here
\begin{tabular}{@{}ccc|cccccc@{}}
\toprule
\textbf{Model} & \textbf{Technique} & \textbf{Pruned} & \textbf{Piqa} & \textbf{Hellaswag} & \textbf{Winogrande} & \textbf{Arc Easy} & \textbf{Wikitextv2} & \textbf{Average} \\ 
& & & Acc. ($\uparrow$) & Acc. ($\uparrow$) & Acc. ($\uparrow$) & Acc. ($\uparrow$) & Perplexity ($\downarrow$) & Acc. ($\uparrow$) \\
\midrule
\multirow{6}{*}{\rotatebox[origin=c]{90}{\textbf{LLaMa-3-8B}}} & \multirow{3}{*}{SliceGPT} & 30\% & 59.3\% & 37.2\% & 56.4\% & 42.9\% & 13.37 & 49.0\% \\
                             &                            & 35\%  & 57.7\% & 34.1\% & 54.3\% & 39.3\% & 16.58 & 46.4\% \\
                             &                            & 40\%  & 57.0\% & 32.4\% & 51.8\% & 35.9\% & 20.69 & 44.3\% \\ \cmidrule{2-9}
                             & \multirow{3}{*}{Dynamic Slicing} & 30\%  & \textbf{60.6\%} & \textbf{38.0\%} & \textbf{57.4\%} & \textbf{43.0\%} & \textbf{13.11} & \textbf{49.8\%} \\
                             &                            & 35\% & \textbf{58.6\%} & \textbf{35.6\%} & \textbf{55.8\%} & \textbf{39.7\%} & \textbf{15.9} & \textbf{47.4\%} \\
                             &                            & 40\% & \textbf{58.1\%} & \textbf{34.0\%} & \textbf{54.6\%} & \textbf{37.1\%} & \textbf{19.34} & \textbf{46.0\%} \\
\cmidrule{2-9}
\multirow{6}{*}{\rotatebox[origin=c]{90}{\textbf{Mistral-7B}}} & \multirow{3}{*}{SliceGPT} & 30\% & 62.6\% & 38.0\% & \textbf{59.7\%} & 51.1\% & 8.87 & 52.9\% \\
                             &                            & 35\% & 58.5\% & \textbf{35.9\%} & \textbf{57.6\%} & 42.8\% & 10.80 & 48.7\% \\
                             &                            & 40\% & 57.1\% & 33.6\% & 54.1\% & 38.2\% & 13.33 & 45.8\% \\ \cmidrule{2-9}
                             & \multirow{3}{*}{Dynamic Slicing} & 30\% & \textbf{62.6\%} & \textbf{38.1\%} & 59.5\% & \textbf{52.3\%} & \textbf{8.81} & \textbf{53.1\%} \\
                             &                            & 35\% & \textbf{59.0\%} & 35.7\% & 57.0\% & \textbf{44.8\%} & \textbf{10.50} & \textbf{49.1\%} \\
                             &                            & 40\% & \textbf{57.8\%} & \textbf{33.6\%} & \textbf{54.2\%} & \textbf{39.6\%} & \textbf{12.76} & \textbf{46.3\%} \\
\bottomrule
\end{tabular}
} % Correctly placed closing brace for \resizebox
\caption{Comparison of our technique in the average setting against the constant slicing proposed by SliceGPT; bold means higher value in comparison}
\label{tab:average_comparison}
\end{table*}

\subsection{Llama3-8B and Mistral-7B in various scenarios}

\begin{figure*}[h]
\centering
\includegraphics[width=1.0\textwidth]{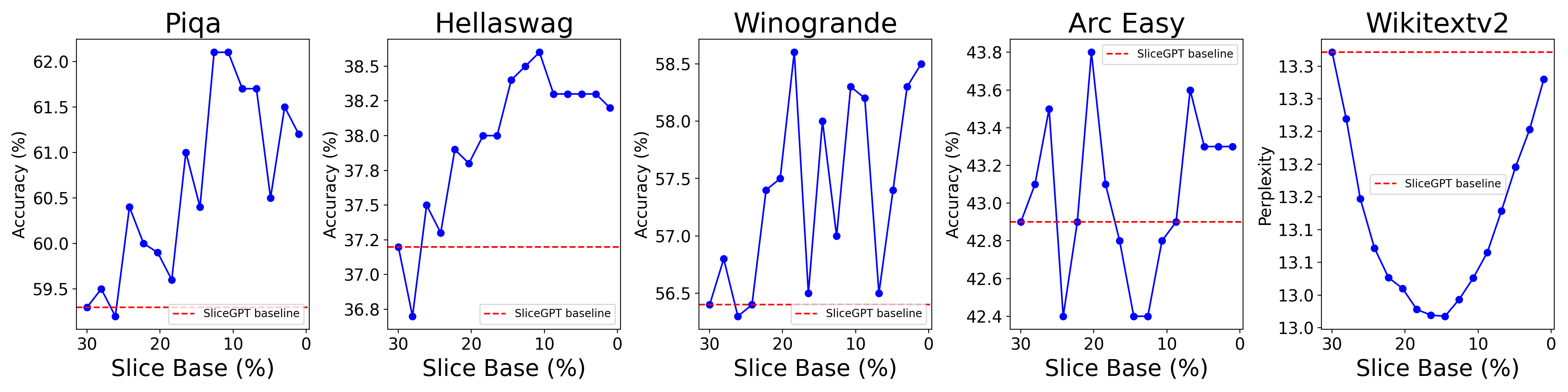}
\caption{Llama3 8B with 30\% of the network sliced on average, the red line is the baseline accuracy achieved by SliceGPT with a constant 30\% slice.}
\label{llama3b30p}
\end{figure*}

\begin{figure*}[h]
\centering
\includegraphics[width=1.0\textwidth]{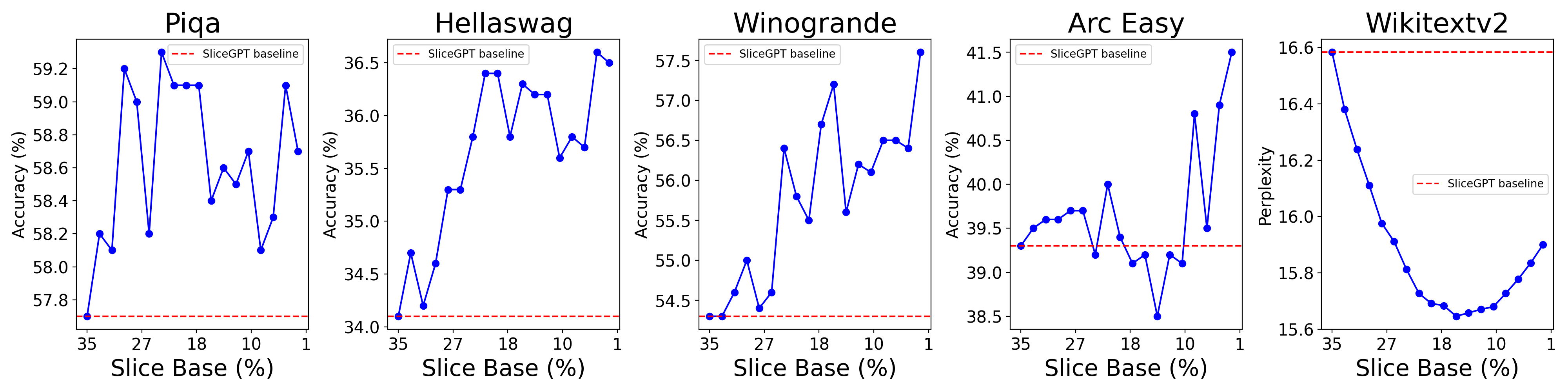}
\caption{Llama3 8B with 35\% of the network sliced on average, the red line is the baseline accuracy achieved by SliceGPT with a constant 35\% slice.}
\label{llama3b35p}
\end{figure*}

\begin{figure*}[h]
\centering
\includegraphics[width=1.0\textwidth]{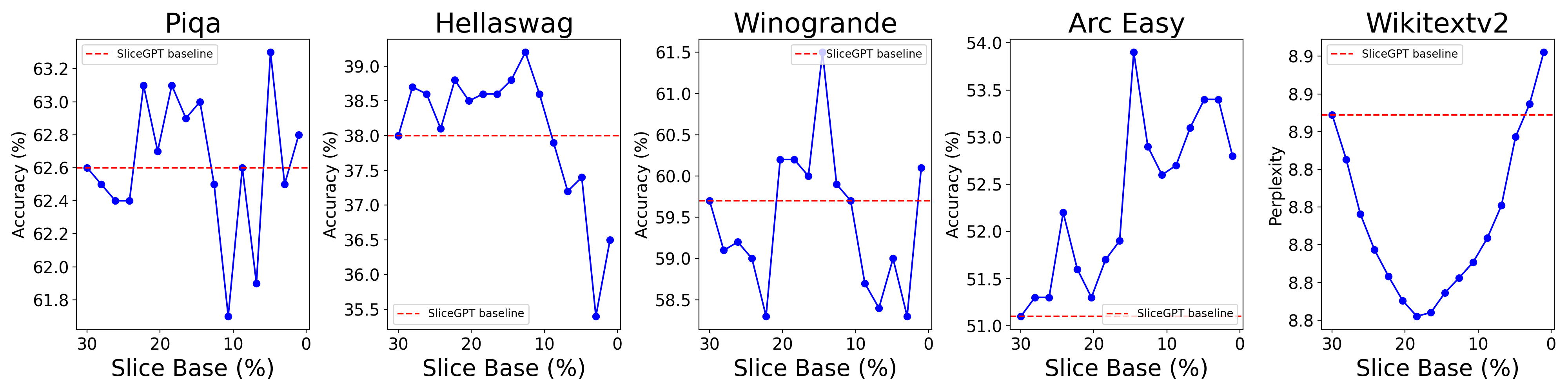}
\caption{Mistral 7B with 30\% of the network sliced on average, the red line is the baseline accuracy achieved by SliceGPT with a constant 30\% slice.}
\label{mistralb30p}
\end{figure*}

\begin{figure*}[h]
\centering
\includegraphics[width=1.0\textwidth]{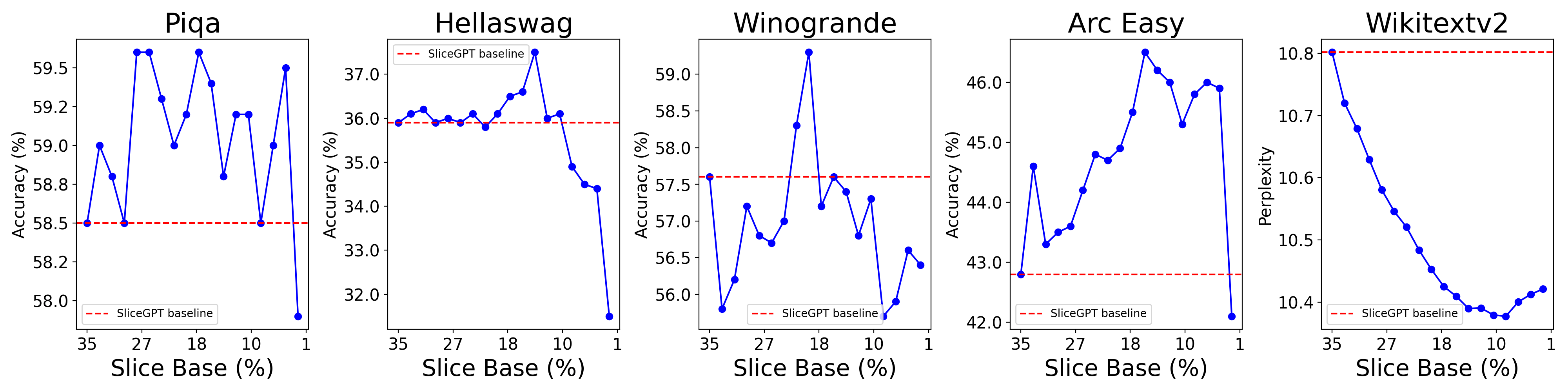}
\caption{Mistral 7B with 35\% of the network sliced on average, the red line is the baseline accuracy achieved by SliceGPT with a constant 35\% slice.}
\label{mistralb35p}
\end{figure*}

\begin{figure*}[h]
\centering
\includegraphics[width=1.0\textwidth]{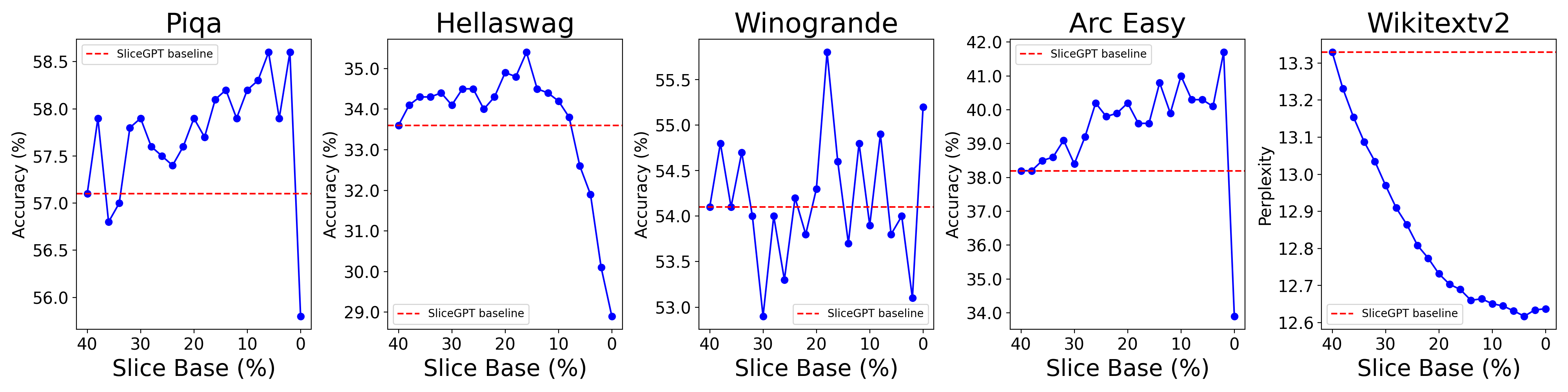}
\caption{Mistral 7B with 40\% of the network sliced on average, the red line is the baseline accuracy achieved by SliceGPT with a constant 40\% slice.}
\label{mistralb40p}
\end{figure*}

Visual representations of our experiments are shown across several figures: Figure \ref{llama3b30p} and Figure \ref{llama3b35p} illustrate the performance of Llama3-8B with 30\% and 35\% of the network sliced, respectively, comparing it to the baseline accuracy achieved by SliceGPT. Similarly, Figures \ref{mistralb30p}, \ref{mistralb35p}, and \ref{mistralb40p} display the results for Mistral-7B with 30\%, 35\%, and 40\% of the network sliced, again benchmarked against SliceGPT's constant slice accuracies. In each figure, the red line denotes the baseline accuracy set by SliceGPT for the respective slicing percentages.

\end{document}